\documentclass[pdflatex,sn-mathphys-num]{sn-jnl}

\usepackage{graphicx}%
\usepackage{multirow}%
\usepackage{amsmath,amssymb,amsfonts}%
\usepackage{amsthm}%
\usepackage{mathrsfs}%
\usepackage[title]{appendix}%
\usepackage{xcolor}%
\usepackage{textcomp}%
\usepackage{manyfoot}%
\usepackage{booktabs}%
\usepackage{algorithm}%
\usepackage{algorithmicx}%
\usepackage{algpseudocode}%
\usepackage{listings}%

\begin{document}

\title[Building Low-Resource African Language Corpora: A Case Study of Kidaw’ida, Kalenjin and Dholuo]{Building Low-Resource African Language Corpora: A Case Study of Kidaw’ida, Kalenjin and Dholuo}


\author*[1]{\fnm{Audrey} \sur{Mbogho}}\email{ambogho@usiu.ac.ke}

\author[1]{\fnm{Quin} \sur{Awuor}}\email{qawuor@usiu.ac.ke}
\equalcont{These authors contributed equally to this work.}

\author[2]{\fnm{Andrew} \sur{Kipkebut}}\email{akipkebut@kabarak.ac.ke}
\equalcont{These authors contributed equally to this work.}

\author[3]{\fnm{Lilian} \sur{Wanzare}}\email{ldwanzare@maseno.ac.ke}
\equalcont{These authors contributed equally to this work.}

\author[3]{\fnm{vivian} \sur{Oloo}}\email{voloo@maseno.ac.ke}
\equalcont{These authors contributed equally to this work.}

\affil*[1]{\orgdiv{School of Science and Technology},
\orgname{USIU-Africa}, \orgaddress{\city{Nairobi}, \country{Kenya}}}

\affil[2]{\orgdiv{Department of Computer Science}, \orgname{Kabarak University}, \orgaddress{\city{Nakuru}, \country{Kenya}}}

\affil[3]{\orgdiv{Department of Computer Science}, \orgname{Maseno University}, \orgaddress{\city{Maseno}, \country{Kenya}}}


\abstract{Natural Language Processing is a crucial frontier in artificial intelligence, with broad applications in many areas, including public health, agriculture, education, and commerce. However, due to the lack of substantial linguistic resources, many African languages remain underrepresented in this digital transformation. This paper presents a case study on the development of linguistic corpora for three under-resourced Kenyan languages, Kidaw'ida, Kalenjin, and Dholuo, with the aim of advancing natural language processing and linguistic research in African communities. Our project, which lasted one year, employed a selective crowd-sourcing methodology to collect text and speech data from native speakers of these languages. Data collection involved (1) recording conversations and translation of the resulting text into Kiswahili, thereby creating parallel corpora, and (2) reading and recording written texts to generate speech corpora. We made these resources freely accessible via open-research platforms, namely Zenodo for the parallel text corpora and Mozilla Common Voice for the speech datasets, thus facilitating ongoing contributions and access for developers to train models and develop Natural Language Processing applications. The project demonstrates how grassroots efforts in corpus building can support the inclusion of African languages in artificial intelligence innovations. In addition to filling resource gaps, these corpora are vital in promoting linguistic diversity and empowering local communities by enabling Natural Language Processing applications tailored to their needs. As African countries like Kenya increasingly embrace digital transformation, developing indigenous language resources becomes essential for inclusive growth. We encourage continued collaboration from native speakers and developers to expand and utilize these corpora.}




\maketitle

\section{Introduction}\label{sec1}

Natural Language Processing (NLP) has become a vital component of modern artificial intelligence (AI), shaping various sectors from healthcare to commerce. In recent years, advances in hardware, sophisticated algorithms, and the availability of large text data sets have enabled NLP systems to deliver unprecedented capabilities, transforming many aspects of human activity \citep{bender2021dangers}. Generative AI tools, such as ChatGPT, exemplify the reach and impact of NLP innovations. However, these advances have exacerbated the global digital divide, particularly in linguistic terms. Many African languages, including those indigenous to Kenya, remain excluded from these developments, limiting access to critical information and technological benefits for speakers of these languages \citep{nekoto2020participatory}.

The linguistic divide is rooted in historical inequalities, mainly from colonialism, which imposed foreign languages like English, French, and Portuguese as the official media of communication in Africa. These languages dominate education, governance, and technology, often at the expense of indigenous languages \citep{bamgbose2011african}. This legacy has led to the marginalization of the majority of the African population, who do not speak these colonial languages fluently, restricting their access to essential information and technological tools developed in those languages \citep{nduati2016post}. During the COVID-19 pandemic, for example, critical public health information in Kenya was predominantly disseminated in English and Kiswahili, excluding speakers of indigenous languages, especially in rural areas.

This linguistic exclusion underscores the urgent need for African languages to be integrated into AI and NLP technologies. Building high-quality linguistic corpora for underrepresented languages is crucial to enable language technologies such as machine translation, speech recognition, and speech synthesis for African languages. In this context, our project focused on developing language corpora for three Kenyan languages: Kidaw'ida, Kalenjin, and Dholuo, by collecting text and speech data and translating the text data into Kiswahili to generate parallel corpora. These resources are intended to facilitate the development of NLP applications that cater to the needs of African communities, thus promoting linguistic diversity in AI development.

\begin{figure}[h]
    \centering
    \includegraphics[width=0.75\textwidth]{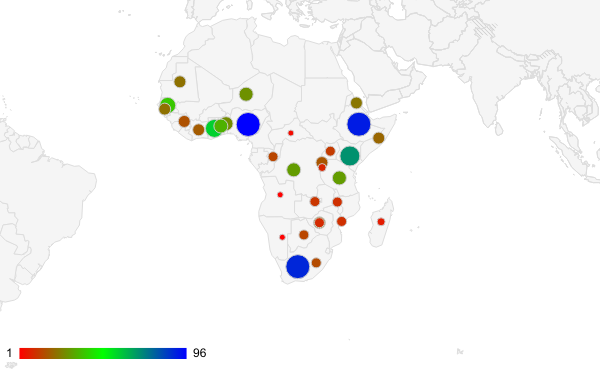}
    \caption{Distribution of NLP activity in Africa based on the number of NLP terms appearing in relation to each country in public sources}
    \label{fig:alrt}
\end{figure}

In the last five years, there has been a surge in NLP activity for African languages. There is a need to get an overall picture of what is being done and where. A search for “African language NLP” reveals a concentration of activity in only a handful of countries, among them South Africa, Nigeria, Ethiopia, Kenya, and Ghana. This state of affairs is depicted in Figure \ref{fig:alrt} based on the number of mentions of specific African languages in publicly available Internet sources. While our project focused on only three Kenyan languages, one of the aims of this paper is to serve as a call to action to initiate similar projects across the African continent and to ensure no language is left behind.

This research aimed to answer the following questions:

\begin{enumerate}
    \item What methods are effective in acquiring data for building language corpora for low-resource languages?
    \item What can motivate members of the target language communities to contribute to language data?
    \item What methods can be applied to ensure the quality of the language data collected?
\end{enumerate}

\section{Background on Kidaw’ida, Kalenjin and Dholuo}\label{sec2}

In this section, we provide background information on each language and describe any language data available for it. We also describe the status of the language insofar as NLP resources are concerned.

\subsection{Kidaw’ida (ISO 639-3 Code: dav)}\label{subsec1}

Kidaw’ida is a Bantu language spoken in Taita Taveta County in Southeastern Kenya by approximately half a million people. Apart from the Bible, and the Anglican Hymn Book, few publications in Kidaw’ida can be found. Frank Mcharo and Peter Bostock published, in Kidaw’ida, a small 71-page book on the customs and traditions of the W’adaw’ida \citep{mcharo1995mizango}. This is a copyrighted resource and can only be used with permission. Furthermore, it is only available in print. Although online articles, social media posts and self-published story books written in Kidaw’ida are known to exist, they are hard to discover and consolidate. Indeed, social media platforms like Facebook, X and WhatsApp could be a rich source of Kidaw’ida text, but to extract it from the mixture of languages used on such platforms presents a significant challenge.

Kidaw’ida faces the threat of glottophagy from the more dominant Kiswahili due to their close geographical proximity and the intermingling of speakers of different  Kenyan languages. Glottophagy is a term coined by the French linguist Louis-Jean Calvet \cite{louis1974linguistique} to describe the ``eating'' of a language by a more dominant one. It perfectly captures what is arguably the most significant current threat to Kidaw’ida. This threat is witnessed in real time as Kiswahili words are rapidly replacing Kidaw’ida words in day to day conversations. Under normal circumstances, borrowing of words is a boon to a language as it gives the language words that it lacks, thus strengthening the expressive power of the language. But the replacement of words when not necessary is very concerning as it could mean the erosion of the culture and the loss of crucial indigenous knowledge. This is not to deny the inevitable evolution of all languages, but rather to raise an alarm about the rapidity of the change affecting languages like Kidaw’ida that are spoken by a small population and that are threatened by a dominant language.

\subsection{Status of Kidaw’ida in NLP}\label{subsec2}

To our knowledge, no other project has been carried out to develop NLP resources for Kidaw’ida. A possible reason for this is the small size and marginalization of the Kidaw’ida speaking population. There has been some interest in Kidaw’ida from linguists and anthropologists, for example, \cite{sakamoto1986social}, but not from NLP researchers. Another important thing to note is that low-resource language research will typically be spearheaded by speakers of that language, or at least involve them in significant roles. The small population of Kidaw’ida speakers makes it less likely to find among them a researcher with the relevant training and the right level of awareness to initiate NLP work. This project, while primarily aiming to generate digital resources for Kidaw’ida for use in NLP, is critical for stemming the rapid decline of the language. It is not far-fetched to imagine that a language that is not useful in the present digital age will soon fall silent.

Mobilising community involvement for the collection of Kidaw’ida language data is crucial. At the same time, we found that it is not easy to get community members involved. There is initial excitement but this does not necessarily translate to a willingness to participate long term. This challenge requires sustained community engagement to overcome, including reasonable remuneration for participation in language projects.

\subsection{Kalenjin (ISO 639-3 Code: kln)}\label{subsec3}

Kalenjin is a Nilotic language, belonging to the Eastern Nilotic branch of the Nilo-Saharan language family. It is primarily spoken by the Kalenjin people, who predominantly reside in the Rift Valley region of Kenya, particularly in counties such as Uasin Gishu, Elgeyo-Marakwet, Bomet, Baringo, Nandi, and Kericho. The language is also spoken by smaller populations in parts of Uganda and Tanzania \cite{naibei2018comparative}. The Kalenjin people are part of the larger Nilotic group that migrated from the Nile Valley to the East African Great Lakes region thousands of years ago \cite{Chelimo2016PreColonialPO}. The migration process, which began around the fifteenth century, saw the Kalenjin spread across the Rift Valley and into the surrounding areas. As a member of the Eastern Nilotic group, Kalenjin shares linguistic features with other languages such as Turkana, Pokot, and Maasai, all of which are spoken by ethnic groups in the same region \cite{naibei2018comparative}. Kalenjin holds significant importance in the cultural and social life of its speakers. It is used for daily communication, in traditional ceremonies, and in storytelling, with oral literature being a vital part of Kalenjin cultural identity. However, similar to many African languages, Kalenjin's use in formal domains like education, governance, and media is limited. In Kenya, English and Kiswahili dominate in these sectors, relegating Kalenjin primarily to informal communication \cite{ogot2002historical}. Kalenjin, like other indigenous languages in Kenya, faces the challenge of language shift, where younger generations tend to prefer English and Kiswahili for social mobility and economic opportunities. Despite this, the language remains a central part of Kalenjin identity and continues to be passed down through generations, particularly in rural areas.

\subsection{Status of Kalenjin in NLP}\label{subsec4}

Kalenjin, similar to other languages in Kenya, is under-resourced in terms of Natural Language Processing (NLP) capabilities. The lack of comprehensive linguistic data, such as text corpora, speech recordings, and annotated resources, hampers the development of effective NLP tools for the language. This scarcity places Kalenjin among the many African languages that are classified as ``low-resource languages'' within the NLP field. Without sufficient linguistic data, it becomes challenging to develop applications like speech recognition, machine translation, or language generation models that can support Kalenjin speakers in the digital world. In the face of rapid technological advancement and the increasing global dominance of English and Kiswahili in digital platforms, Kalenjin risks being digitally marginalised. Without substantial efforts to build linguistic resources, including corpora and NLP tools, Kalenjin could become excluded from AI-powered communication technologies. This exclusion would further deepen the divide between speakers of major global languages and those of indigenous languages, leaving Kalenjin speakers without adequate access to digital services, content, and resources that are increasingly available to speakers of dominant languages \cite{bamgbose2011african}.

However, there is potential for revitalising Kalenjin in the digital space. Just as initiatives in other African languages are showing success in developing NLP tools, similar efforts can be made for Kalenjin. By collecting text and speech data, building annotated corpora, and fostering community involvement, there is an opportunity to ensure that Kalenjin is represented in the digital landscape. These steps would not only preserve the language for future generations but also empower its speakers to take part in technological advancements and digital platforms in a meaningful way. To address these challenges, various stakeholders, including linguists, technologists, government institutions, and community organisations, must collaborate to create a sustainable ecosystem for Kalenjin language preservation and digital integration. Such efforts would help ensure that Kalenjin retains its cultural and linguistic relevance in an increasingly interconnected world. 

\subsection{Dholuo (ISO 639-3 Code: luo)}\label{subsec5}

Dholuo is a Nilotic language belonging to the Western Nilotic branch of the Nilo-Saharan language family. It is primarily spoken by the Luo ethnic group, which resides predominantly in the western parts of Kenya, along the shores of Lake Victoria, and parts of Tanzania and Uganda \cite{omulo2018survey}. The language traces its origins to the migration of Nilotic peoples from southern Sudan, who gradually settled in the Great Lakes region of East Africa around the 15th century \cite{heine2000african}. As a Western Nilotic language, Dholuo shares linguistic similarities with other languages in that family, such as Acholi and Lango spoken in Uganda and South Sudan.

Dholuo plays an important role in the cultural and social life of the Luo people, serving as a medium of communication in daily life, traditional practices, and artistic expressions such as music and oral literature. However, its usage is restricted to informal settings, as English and Kiswahili dominate formal domains such as education, governance, and commerce in Kenya \cite{nduati2016post}.

\subsection{Status of Dholuo in NLP}\label{subsec6}

Linguistic data for Dholuo, including text and speech corpora, are limited in availability, which is crucial for developing NLP applications like machine translation, speech recognition, and language generation systems. This paucity of resources places Dholuo in the category of low-resource languages, meaning it needs more extensive datasets and digital tools for robust AI and NLP model development \cite{nekoto2020participatory}. This lack of linguistic resources compounds the risk of Dholuo marginalisation in the digital age. Without concerted efforts to build corpora and develop NLP tools, Dholuo faces the threat of digital extinction, where its speakers are excluded from technological advancements, particularly in AI-powered communication platforms. This further exacerbates the language's vulnerability as it struggles to maintain relevance in a globalised world where dominant languages continue to shape technological access and innovation \cite{bamgbose2011african}.

To address this, efforts are needed to develop NLP resources for Dholuo by collecting text and speech data, creating annotated corpora, and encouraging community participation in language preservation projects. Such interventions are necessary for Dholuo to avoid becoming even more marginalised, both linguistically and digitally.

\section{Related Work}\label{sec3}

Many African languages are primarily oral, with limited written materials and even less electronically available. This means that standard approaches that harvest online data, such as web crawling, are excluded in many cases. Therefore, for many African languages, researchers must rely on other methods to collect data, such as fieldwork, community engagement, and collaborations with native speakers. By actively seeking out and collecting data through these means, researchers can ensure that their work is inclusive and representative of the diverse linguistic landscape of Africa. This section reviews a few examples of other researchers' approaches to building African language corpora from which we drew inspiration for our own project.

Kencorpus \cite{wanjawa-etal-2023-kencorpus} is a corpus of Swahili, Dholuo, and Luhya, three of the more commonly spoken languages in Kenya. The corpus includes text documents, voice files, and a question-answering dataset. Data collection was done by researchers who were deployed to communities, schools and media houses. Kencorpus collected 4,442 text sentences and 177 hours of recorded speech. From the raw data, the project annotated Dholuo and Luhya texts with part of speech tags and created question-answer pairs for the Kiswahili corpus. In addition, a parallel corpus was created by translating the Dholuo and Luhya datasets into Kiswahili.

Adelani et al. \cite{adelani2021masakhaner} used online news sites to develop named entity recognition (NER) datasets for ten African languages, namely, Amharic, Hausa, Igbo, Kinyarwanda, Luganda, Luo, Nigerian-Pidgin, Swahili, Wolof and Yorùbá. The data consisted of about 30,000 sentences in total, annotated for the NER task. The population sizes of the speakers of the ten languages range from 4 million for Dholuo to 98 million for Swahili. Volunteers annotated the raw data with entity labels for person, location, organisation and date. The annotators were part of the Masakhane community and were not paid, but were included as co-authors on the paper. Multiple annotators of the same entities provided redundancy and quality assurance. This approach offers a viable alternative to the more common paid or unpaid crowdsourcing approach. Volunteers who are part of a community such as Masakhane that has been set up for language work are already motivated and knowledgeable and may be better prepared to participate.

Nakatumba-Nabende et al. \cite{nakatumba2024building} developed text and speech resources for four Ugandan languages and Swahili. The four Ugandan languages are Luganda, Runyankore-Rukinga, Lumasaba, and Acholi. Some of the text data was created by translating English text into the five African languages. The English text was obtained from different published sources, including English Wikipedia, social media, online local newspapers, story books, novels and human rights charters. This process, as well as generating data for the African languages, also generated parallel corpora for English and the African languages. Additional text data was obtained by recruiting native speakers through crowdsourcing to write sentences and review them for quality. Due to lack of access to computers, some of the contributors wrote their sentences down by hand, and these had to subsequently be typed. Text sentences were later read by native speakers, again through crowdsourcing, and recorded to generate speech corpora. Five parallel corpora were created consisting of 40,000 sentence pairs each. A total of  sentences of monolingual data for five languages was collected. The project also recorded  582 hours of Luganda speech and 1100 hours of Swahili speech.

Ogayo et al. \cite{ogayo2022building} built a speech synthesis dataset for eleven African languages, namely, Dhouo, Lingala, Kikuyu, Yoruba, Hausa, Luganda, Ibibio, Kiswahili, Wolof, Fongbe and Suba. The data consisted of a total of 65,537 utterances distributed unevenly across languages, ranging from 125 to 11,971 utterances. These utterances corresponded to a total of 113.84 hours of recordings, ranging from 0.33 to 24.82 hours. Religious texts and other online and print sources were used as data sources. The data was supplemented with contributions from recruited participants. To help encourage community participation, comprehensive guidance on data collection methods and data licensing was provided. Because of the quality of speech required for speech synthesis, voice talents were carefully vetted and were remunerated in cash and kind.

The examples discussed in this section give an idea of the kinds of activities that are going on in corpus building for African language NLP, and suggest approaches that have been shown to work well and that others can emulate. It is essential to note that these examples focus on specific regions and languages and only partially represent Africa's linguistic diversity. They also only offer a glimpse into the vast amount of NLP work (including corpus building) that is currently going on in Africa. As momentum in African language technologies picks up, researchers need to be intentional in ensuring that no language is left behind and that global inequalities are not replicated locally.

As these examples show, current African language projects, regardless of the methods used, only generate modest amounts of data, typically in the order of tens of thousands of tokens or hundreds of hours of recorded speech. On the other hand, the latest NLP advancements are in large language models (LLMs), which require massive datasets, with Llama 2 from Meta having been trained on 2 trillion tokens \cite{Touvron2023Llama2O}. OpenAI’s latest LLM has been trained on even more data than this (the specifics have not been officially published). It is therefore important that the methods employed for data collection are scalable and sustainable to support more expansive research endeavours across the continent. This is a significant challenge that requires innovative solutions.

\section{Our Methodology}\label{sec4}

Our main objective was to collect text and speech data for three indigenous Kenyan languages, namely, Kidaw’ida, Kalenjin and Dholuo. These languages were chosen because they are the home languages of the research team. The project proposed to source the data from members of the language communities through crowdsourcing. The data collection was realised through a grant from the Lacuna Fund, which was used to pay small stipends to compensate language data contributors for their time, effort and the knowledge shared.

\subsection{Text Data Collection}\label{subsec1}

Two approaches were used for generating text data:
\begin{enumerate}
\item Contributors wrote down sentences in the three languages covered by this project. These could be made up sentences or sentences from materials that are in the public domain. Some contributors wanted to use the bible but most local language bibles in Kenya are copyrighted by the Bible Society of Kenya and the British and Foreign Bible Society. However we advised that they could use such copyrighted materials for inspiration. For example, a Bible story could remind them of something else they could write about. The issue of cultural appropriateness of data is often cited as militating against using texts of foreign origin, but we propose that such text can catalyse ideas.
\item Contributors recorded conversations in the three languages, which were later transcribed. The participants in the conversations were informed about the recording in advance and were asked to sign a consent form. During transcription, words that were in a language different from the target language were replaced as code-switched speech was outside the scope of the project.
\end{enumerate}

The sentences collected were translated into Kiswahili to generate three parallel corpora. Microsoft Excel and Google Sheets were used to compose and translate sentences. The resulting spreadsheets were stored on Google Drive and GitHub while the project was in progress. 

Quality Assurance was achieved by identifying contributors with high levels of language proficiency who were designated the role of Data Collection Leads (DCLs) and were paid a monthly salary. The DCLs identified language contributors they knew were qualified and recommended them for recruitment. DCLs also contributed data and checked the contributors’ data for correct spelling, grammar, fluency, and proper translation into Kiswahili. The DCLs corrected any errors they found in the data.

To boost female representation in NLP work, 7 of the DCLs were female and 5 were male, and  5 of the researchers were female and 1 was male. Women were also well represented among the contributors as can be seen in Table \ref{tab1} below. It is important to have gender balance in language projects to ensure applications work for everyone and also to avoid bias, for example, in generative AI.

\subsection{Voice Data Collection}\label{subsec2}

To allow our geographically distributed contributors to make voice data contributions easily, we determined that an open-access online platform would be ideal. Quite surprisingly, we were only able to find two options, namely, Living Dictionaries (\url{https://livingdictionaries.app}) and Mozilla Common Voice (\url{https://commonvoice.mozilla.org}).

Initially, it seemed that Living Dictionaries would be easier to use. Mozilla Common Voice involves a rather steep on-ramp, which we explain below. Living Dictionaries, although easy to get started on, turned out not to be fit for purpose as the main aim of that platform is language documentation and it is tailored for  linguistic work rather than NLP projects. For example, when recording voice for NLP, it is required that the same phrase be read by multiple people. This is not well supported on Living Dictionaries as the interest there is in capturing how something is said and one person saying it suffices. However, in NLP, and more specifically in automatic speech recognition (ASR), the different characteristics of people's voices while saying the same phrase must be captured, so that applications built on this data can work for everyone.

Thus, in the end we had to go with Mozilla Common Voice. The first step in order to get onto Mozilla Common Voice is to make a language request. Once this request was approved for each of our three languages, we had to localize the pages associated with our languages using a platform provided by Mozilla known as Pontoon (\url{https://pontoon.mozilla.org}). This entails translating over one thousand technical strings into each language, a task that requires both technical expertise and language proficiency, including the ability to invent appropriate words for technical terms that do not yet exist in the language, such as “database”, “download” or “speech recognition.” This was quite time-consuming and was a required step before the language was launched on Mozilla Common Voice. Once launched, we were now able to upload the text sentences we had collected earlier, and once enough sentences were available on the platform, we were able to start reading and recording. Our DCLs recruited more people to read and record, to ensure the diversity of voices required in speech data.

The Kidaw’ida, Kalenjin and Dholuo text sentences were uploaded onto Mozilla Common Voice to allow members of the three language communities to contribute their voices. 

\section{Results}\label{sec5}

We collected 30,000 text sentences for each of the three languages and had contributors proficient in both their mother tongue and  Kiswahili provide translations. This resulted in the creation of three parallel corpora: Kidaw’ida - Kiswahili, Kalenjin - Kiswahili and Dholuo - Kiswahili. These parallel corpora are freely available for download from \url{https://zenodo.org/records/13355021} (Mbogho et al., 2024). The same repository is on Github where it can be updated and reuploaded to Zenodo. This is important not only for the continual expansion of the dataset but also for making any needed corrections and quality improvements.

The collection of voice data is ongoing, but at present the speech data on Mozilla Common Voice is as shown in Table \ref{tab1}.

\begin{table}[h]
\caption{Speech Data for Kidaw’ida, Kalenjin and Dholuo on Mozilla Common Voice}\label{tab1}%
\begin{tabular}{@{}lllll@{}}
\toprule
Language & Hours  & Speakers & Female  & Male\\
\midrule
Kidaw'ida   & 56   & 24  & 60\%  & 40\%  \\
Kalenjin    & 92   & 41 & 70\%   & 30\%  \\
Dholuo    & 120   & 44  & 58\%  & 42\%  \\
\botrule
\end{tabular}
\end{table}

\section{Discussion}\label{sec6}

Selected crowdsourcing was found to be a useful approach to developing corpora for African languages which balances rapid data generation with quality assurance. Although many people are keen to contribute data for the languages to promote and preserve them, funding is essential to motivate broader participation, especially to the level required for generating large datasets within a reasonable amount of time.

Anyone who wishes to download the voice data from Mozilla Common Voice can do so at no cost and without the need for attribution because the license is CC0. Developers are encouraged to take advantage of this unrestricted access to the data to train models and create applications for these three languages. The language communities are encouraged to continue to add to the repositories so that greater accuracies in models trained from the data can be achieved.

\section{Conclusions and Future Work}\label{sec7}

The project ran for a year and generated 30,000 sentences per language. This still needs to be increased to train NLP models effectively. Future work will look at obtaining a more substantial amount of funding to recruit more contributors.  However, the datasets should still be used in their current state to train models in order to establish a baseline that is an indication of the level of accuracy achievable currently. It is also important to document what is possible with small datasets as not all applications require massive amounts of data.

As the amount of data grows and reaches levels that can train accurate models for data-hungry tasks, NLP developers in Kenya need to turn their attention to developing applications with local languages. Areas where local language NLP technology can be beneficial include health, agriculture, transport, education, and commerce, among many others. Successful applications will, in turn, generate more data, leading to improvements in model accuracy in the future.

We have referred to the type of crowd-sourcing we used as ``selective'' because the contributors were known to members of the project team. This was important because language proficiency was a key consideration, and using a “random” crowd could have negatively impacted the quality of the data. However, this approach may not be sustainable for large projects that are looking to get a large amount of data in a short time. In that case, a project has to risk collecting low quality data and then applying quality control measures afterwards.

Licensing issues are a major concern in corpus building for low-resource languages. A greater involvement of intellectual property specialists is essential to ensure that language communities are not disadvantaged but rather stand to benefit from the resulting data and associated applications.

\section*{Acknowledgements}

This work was carried out with support from Lacuna Fund, an initiative co‐founded by The Rockefeller Foundation, Google.org, Canada’s International Development Research Centre, and GIZ on behalf of the German Federal Ministry for Economic Cooperation and Development (BMZ). The authors also wish to thank the Kidaw'ida, Kalenjin and Dholuo language communities for sharing their languages with the world.

\section*{Declarations}

The views expressed herein do not necessarily represent those of Lacuna Fund, its Steering Committee, its funders, or Meridian Institute.

\bibliography{sn-bibliography}


\begin{thebibliography}{17}
\ifx \bisbn   \undefined \def \bisbn  #1{ISBN #1}\fi
\ifx \binits  \undefined \def \binits#1{#1}\fi
\ifx \bauthor  \undefined \def \bauthor#1{#1}\fi
\ifx \batitle  \undefined \def \batitle#1{#1}\fi
\ifx \bjtitle  \undefined \def \bjtitle#1{#1}\fi
\ifx \bvolume  \undefined \def \bvolume#1{\textbf{#1}}\fi
\ifx \byear  \undefined \def \byear#1{#1}\fi
\ifx \bissue  \undefined \def \bissue#1{#1}\fi
\ifx \bfpage  \undefined \def \bfpage#1{#1}\fi
\ifx \blpage  \undefined \def \blpage #1{#1}\fi
\ifx \burl  \undefined \def \burl#1{\textsf{#1}}\fi
\ifx \doiurl  \undefined \def \doiurl#1{\url{https://doi.org/#1}}\fi
\ifx \betal  \undefined \def \betal{\textit{et al.}}\fi
\ifx \binstitute  \undefined \def \binstitute#1{#1}\fi
\ifx \binstitutionaled  \undefined \def \binstitutionaled#1{#1}\fi
\ifx \bctitle  \undefined \def \bctitle#1{#1}\fi
\ifx \beditor  \undefined \def \beditor#1{#1}\fi
\ifx \bpublisher  \undefined \def \bpublisher#1{#1}\fi
\ifx \bbtitle  \undefined \def \bbtitle#1{#1}\fi
\ifx \bedition  \undefined \def \bedition#1{#1}\fi
\ifx \bseriesno  \undefined \def \bseriesno#1{#1}\fi
\ifx \blocation  \undefined \def \blocation#1{#1}\fi
\ifx \bsertitle  \undefined \def \bsertitle#1{#1}\fi
\ifx \bsnm \undefined \def \bsnm#1{#1}\fi
\ifx \bsuffix \undefined \def \bsuffix#1{#1}\fi
\ifx \bparticle \undefined \def \bparticle#1{#1}\fi
\ifx \barticle \undefined \def \barticle#1{#1}\fi
\bibcommenthead
\ifx \bconfdate \undefined \def \bconfdate #1{#1}\fi
\ifx \botherref \undefined \def \botherref #1{#1}\fi
\ifx \url \undefined \def \url#1{\textsf{#1}}\fi
\ifx \bchapter \undefined \def \bchapter#1{#1}\fi
\ifx \bbook \undefined \def \bbook#1{#1}\fi
\ifx \bcomment \undefined \def \bcomment#1{#1}\fi
\ifx \oauthor \undefined \def \oauthor#1{#1}\fi
\ifx \citeauthoryear \undefined \def \citeauthoryear#1{#1}\fi
\ifx \endbibitem  \undefined \def \endbibitem {}\fi
\ifx \bconflocation  \undefined \def \bconflocation#1{#1}\fi
\ifx \arxivurl  \undefined \def \arxivurl#1{\textsf{#1}}\fi
\csname PreBibitemsHook\endcsname

\bibitem[\protect\citeauthoryear{Bender et~al.}{2021}]{bender2021dangers}
\begin{bchapter}
\bauthor{\bsnm{Bender}, \binits{E.M.}},
\bauthor{\bsnm{Gebru}, \binits{T.}},
\bauthor{\bsnm{McMillan-Major}, \binits{A.}},
\bauthor{\bsnm{Shmitchell}, \binits{S.}}:
\bctitle{On the dangers of stochastic parrots: Can language models be too big?}
In: \bbtitle{Proceedings of the 2021 ACM Conference on Fairness, Accountability, and Transparency},
pp. \bfpage{610}--\blpage{623}
(\byear{2021})
\end{bchapter}
\endbibitem

\bibitem[\protect\citeauthoryear{Nekoto et~al.}{2020}]{nekoto2020participatory}
\begin{botherref}
\oauthor{\bsnm{Nekoto}, \binits{W.}},
\oauthor{\bsnm{Marivate}, \binits{V.}},
\oauthor{\bsnm{Matsila}, \binits{T.}},
\oauthor{\bsnm{Fasubaa}, \binits{T.}},
\oauthor{\bsnm{Kolawole}, \binits{T.}},
\oauthor{\bsnm{Fagbohungbe}, \binits{T.}},
\oauthor{\bsnm{Akinola}, \binits{S.O.}},
\oauthor{\bsnm{Muhammad}, \binits{S.H.}},
\oauthor{\bsnm{Kabongo}, \binits{S.}},
\oauthor{\bsnm{Osei}, \binits{S.}}, et al.:
Participatory research for low-resourced machine translation: A case study in african languages.
arXiv preprint arXiv:2010.02353
(2020)
\end{botherref}
\endbibitem

\bibitem[\protect\citeauthoryear{Bamgbose}{2011}]{bamgbose2011african}
\begin{bchapter}
\bauthor{\bsnm{Bamgbose}, \binits{A.}}:
\bctitle{African languages today: The challenge of and prospects for empowerment under globalization}.
In: \bbtitle{Selected Proceedings of the 40th Annual Conference on African Linguistics},
pp. \bfpage{1}--\blpage{14}
(\byear{2011}).
\bcomment{Cascadilla Proceedings Project Somerville}
\end{bchapter}
\endbibitem

\bibitem[\protect\citeauthoryear{Nduati}{2016}]{nduati2016post}
\begin{botherref}
\oauthor{\bsnm{Nduati}, \binits{R.N.}}:
The post-colonial language and identity experiences of transnational kenyan teachers in us universities.
PhD thesis,
Syracuse University
(2016)
\end{botherref}
\endbibitem

\bibitem[\protect\citeauthoryear{Mcharo}{1995}]{mcharo1995mizango}
\begin{bbook}
\bauthor{\bsnm{Mcharo}, \binits{A.F.}}:
\bbtitle{Mizango na Maza Ra Kidawida Ra Kufuma Kokala}.
\bpublisher{Peter Bostock},
\blocation{Kenya}
(\byear{1995}).
\burl{https://books.google.co.ke/books?id=0zOBHAAACAAJ}
\end{bbook}
\endbibitem

\bibitem[\protect\citeauthoryear{Calvet}{1974}]{louis1974linguistique}
\begin{botherref}
\oauthor{\bsnm{Calvet}, \binits{L.-J.}}:
Linguistique et colonialisme.
Petit trait{\'e} de glottophagie, Paris, Payot
(1974)
\end{botherref}
\endbibitem

\bibitem[\protect\citeauthoryear{Sakamoto}{1986}]{sakamoto1986social}
\begin{barticle}
\bauthor{\bsnm{Sakamoto}, \binits{K.}}:
\batitle{Social organization and ritual among the taita of kenya the process of the social change from kichuku type to muzi type}.
\bjtitle{Journal of African Studies}
\bvolume{1986}(\bissue{28}),
\bfpage{27}--\blpage{47}
(\byear{1986})
\end{barticle}
\endbibitem

\bibitem[\protect\citeauthoryear{Naibei~Faith and Lwangale}{2018}]{naibei2018comparative}
\begin{barticle}
\bauthor{\bsnm{Naibei~Faith}, \binits{K.}},
\bauthor{\bsnm{Lwangale}, \binits{D.}}:
\batitle{A comparative study of the kalenjin dialects}.
\bjtitle{International Journal of Academic Research in Business and Social Sciences}
\bvolume{8}(\bissue{8}),
\bfpage{476}--\blpage{503}
(\byear{2018})
\end{barticle}
\endbibitem

\bibitem[\protect\citeauthoryear{Chelimo and Chelelgo}{2016}]{Chelimo2016PreColonialPO}
\begin{botherref}
\oauthor{\bsnm{Chelimo}, \binits{F.J.}},
\oauthor{\bsnm{Chelelgo}, \binits{K.}}:
Pre-colonial political organization of the kalenjin of kenya: An overview.
International journal of innovative research and development
\textbf{5}
(2016)
\end{botherref}
\endbibitem

\bibitem[\protect\citeauthoryear{Ogot}{2002}]{ogot2002historical}
\begin{barticle}
\bauthor{\bsnm{Ogot}, \binits{B.A.}}:
\batitle{Historical portrait of western kenya up to 1895 bethwell a. ogot}.
\bjtitle{Historical Studies and Social Change in Western Kenya: Essays in Memory of Professor Gideon S. Were}
\bvolume{13}(\bissue{4}),
\bfpage{99}--\blpage{120}
(\byear{2002})
\end{barticle}
\endbibitem

\bibitem[\protect\citeauthoryear{Omulo and Williams}{2018}]{omulo2018survey}
\begin{barticle}
\bauthor{\bsnm{Omulo}, \binits{A.G.}},
\bauthor{\bsnm{Williams}, \binits{J.J.}}:
\batitle{A survey of the influence of ‘ethnicity’, in african governance, with special reference to its impact in kenya vis-{\`a}-vis its luo community}.
\bjtitle{African Identities}
\bvolume{16}(\bissue{1}),
\bfpage{87}--\blpage{102}
(\byear{2018})
\end{barticle}
\endbibitem

\bibitem[\protect\citeauthoryear{Heine and Nurse}{2000}]{heine2000african}
\begin{bbook}
\bauthor{\bsnm{Heine}, \binits{B.}},
\bauthor{\bsnm{Nurse}, \binits{D.}}:
\bbtitle{African Languages: An Introduction}.
\bpublisher{Cambridge University Press},
\blocation{United Kingdom}
(\byear{2000})
\end{bbook}
\endbibitem

\bibitem[\protect\citeauthoryear{Wanjawa et~al.}{2023}]{wanjawa-etal-2023-kencorpus}
\begin{bchapter}
\bauthor{\bsnm{Wanjawa}, \binits{B.}},
\bauthor{\bsnm{Wanzare}, \binits{L.}},
\bauthor{\bsnm{Indede}, \binits{F.}},
\bauthor{\bsnm{McOnyango}, \binits{O.}},
\bauthor{\bsnm{Ombui}, \binits{E.}},
\bauthor{\bsnm{Muchemi}, \binits{L.}}:
\bctitle{Kencorpus: A kenyan language corpus of {S}wahili, dholuo and luhya for natural language processing tasks}.
In: \beditor{\bsnm{Wartena}, \binits{C.}} (ed.)
\bbtitle{Journal for Language Technology and Computational Linguistics, Vol. 36 No. 2},
pp. \bfpage{1}--\blpage{27}.
\bpublisher{German Society for Computational Linguistics and Language Technology},
\blocation{unknown}
(\byear{2023}).
\doiurl{10.21248/jlcl.36.2023.243}
\end{bchapter}
\endbibitem

\bibitem[\protect\citeauthoryear{Adelani et~al.}{2021}]{adelani2021masakhaner}
\begin{barticle}
\bauthor{\bsnm{Adelani}, \binits{D.I.}},
\bauthor{\bsnm{Abbott}, \binits{J.}},
\bauthor{\bsnm{Neubig}, \binits{G.}},
\bauthor{\bsnm{D’souza}, \binits{D.}},
\bauthor{\bsnm{Kreutzer}, \binits{J.}},
\bauthor{\bsnm{Lignos}, \binits{C.}},
\bauthor{\bsnm{Palen-Michel}, \binits{C.}},
\bauthor{\bsnm{Buzaaba}, \binits{H.}},
\bauthor{\bsnm{Rijhwani}, \binits{S.}},
\bauthor{\bsnm{Ruder}, \binits{S.}}, \betal:
\batitle{Masakhaner: Named entity recognition for african languages}.
\bjtitle{Transactions of the Association for Computational Linguistics}
\bvolume{9},
\bfpage{1116}--\blpage{1131}
(\byear{2021})
\end{barticle}
\endbibitem

\bibitem[\protect\citeauthoryear{Nakatumba-Nabende et~al.}{2024}]{nakatumba2024building}
\begin{barticle}
\bauthor{\bsnm{Nakatumba-Nabende}, \binits{J.}},
\bauthor{\bsnm{Babirye}, \binits{C.}},
\bauthor{\bsnm{Nabende}, \binits{P.}},
\bauthor{\bsnm{Tusubira}, \binits{J.F.}},
\bauthor{\bsnm{Mukiibi}, \binits{J.}},
\bauthor{\bsnm{Wairagala}, \binits{E.P.}},
\bauthor{\bsnm{Mutebi}, \binits{C.}},
\bauthor{\bsnm{Bateesa}, \binits{T.S.}},
\bauthor{\bsnm{Nahabwe}, \binits{A.}},
\bauthor{\bsnm{Tusiime}, \binits{H.}}, \betal:
\batitle{Building text and speech benchmark datasets and models for low-resourced east african languages: Experiences and lessons}.
\bjtitle{Applied AI Letters}
\bvolume{5}(\bissue{2}),
\bfpage{92}
(\byear{2024})
\end{barticle}
\endbibitem

\bibitem[\protect\citeauthoryear{Ogayo et~al.}{2022}]{ogayo2022building}
\begin{bchapter}
\bauthor{\bsnm{Ogayo}, \binits{P.}},
\bauthor{\bsnm{Neubig}, \binits{G.}},
\bauthor{\bsnm{Black}, \binits{A.W.}}:
\bctitle{Building tts systems for low resource languages under resource constraints}.
In: \bbtitle{Proc. S4SG 2022}
(\byear{2022})
\end{bchapter}
\endbibitem

\bibitem[\protect\citeauthoryear{Touvron et~al.}{2023}]{Touvron2023Llama2O}
\begin{botherref}
\oauthor{\bsnm{Touvron}, \binits{H.}},
\oauthor{\bsnm{Martin}, \binits{L.}},
\oauthor{\bsnm{Stone}, \binits{K.R.}},
\oauthor{\bsnm{Albert}, \binits{P.}},
\oauthor{\bsnm{Almahairi}, \binits{A.}},
\oauthor{\bsnm{Babaei}, \binits{Y.}},
\oauthor{\bsnm{Bashlykov}, \binits{N.}},
\oauthor{\bsnm{Batra}, \binits{S.}},
\oauthor{\bsnm{Bhargava}, \binits{P.}}, et al.:
Llama 2: Open foundation and fine-tuned chat models.
ArXiv
\textbf{abs/2307.09288}
(2023)
\end{botherref}
\endbibitem

\end{thebibliography}

\end{document}